\date{July 29, 2025}
\PassOptionsToPackage{hidelinks}{hyperref}

\PassOptionsToPackage{table,xcdraw}{xcolor}

\documentclass{article}

\usepackage[colorinlistoftodos]{todonotes}
\usepackage{booktabs}
\usepackage{float} 
\usepackage{enumitem}
\usepackage{array}
\usepackage{tabularx}
\usepackage{amssymb}
\usepackage{graphicx}  
\usepackage{url}
\usepackage{amsmath}
\usepackage{hyperref}
\usepackage{adjustbox}
\usepackage{times}
\usepackage[margin=1in]{geometry}
\usepackage[table,xcdraw]{xcolor}

\hypersetup{
  pdftitle={Bridging Clear and Adverse Driving Conditions},
  pdfauthor={Yoel Shapiro, Yahia Showgan, Mullick Koustav},
  pdfkeywords={domain adaptation, diffusion, GANs, autonomous driving, semantic segmentation}
}

\title{Bridging Clear and Adverse Driving Conditions: Domain Adaptation with Simulation, Diffusion, and GANs}

\author{
  Yoel Shapiro \\
  \texttt{Yoel.Shapiro@il.bosch.com}
  \and
  Yahia Showgan \\
  \texttt{YahiaShowgan@gmail.com}
  \and
  Mullick Koustav \\
  \texttt{Koustav.Mullick@in.bosch.com}
}

\date{Bosch Center for Artificial Intelligence, Robert Bosch GmbH\\[1ex]July 29, 2025}

\begin{document}

\maketitle
\begin{abstract}
Autonomous Driving (AD) systems exhibit markedly degraded performance under adverse environmental conditions, such as low illumination and precipitation.
The underrepresentation of adverse conditions in AD datasets makes it challenging to address this deficiency.
To circumvent the prohibitive cost of acquiring and annotating real adverse‑weather data, we propose a novel Domain Adaptation (DA) pipeline that transforms clear-weather images into fog, rain, snow, and nighttime images.
Here, we systematically develop and evaluate several novel data-generation pipelines, including\\ simulation-only, GAN-based, and hybrid diffusion-GAN approaches, to synthesize photorealistic adverse images from labelled clear images.
We leverage an existing DA GAN, extend it to support auxiliary inputs, and develop a novel training recipe that leverages both simulated and real images.
The simulated images facilitate exact supervision by providing perfectly matched image pairs, while the real images help bridge the simulation-to-real (sim2real) gap.
We further introduce a method to mitigate hallucinations and artifacts in Stable-Diffusion Image-to-Image (img2img) outputs by blending them adaptively with their progenitor images.
We finetune downstream models on our synthetic data and evaluate them on the Adverse Conditions Dataset with Correspondences (ACDC).
We achieve 1.85\% overall improvement in semantic segmentation, and 4.62\% on nighttime, demonstrating the efficacy of our hybrid method for robust AD perception under challenging conditions. 
A live demo showcasing our domain adaptation pipeline and results is available at:
\href{https://yahiashowgan.github.io/bridging-driving-conditions/}{yahiashowgan.github.io/bridging-driving-conditions}.


\end{abstract}

\section{Introduction}
\label{intro}

Adverse weather causes a disproportionate share of accidents (50\% of fatalities occur at night and 14\% in rain\cite{fhwa_weather, destatis2024causes, dougmilnes2024dangerous}). Unfortunately, collecting and labelling such data is costly and unsafe, yielding a persistent clear‑weather bias in existing datasets. Perception failures have been identified as the primary cause in at least half of these incidents. This motivates research into alternative data generation methods, such as simulation, domain adaptation (DA), and generative techniques. Common limitations are reliance on unlabelled target data, heavyweight backbones, or loss of pixel-level consistency. In this work, we propose a target-data-free approach to synthesize high-quality adverse-condition training data using only labelled clear-weather images. Our goal is to bridge the domain gap with minimal supervision using a practical, task-agnostic pipeline.
Unlike prior DA methods that operate only in simulation, we then apply our simulator-trained DA model directly to real clear images from the ACDC-Clear subset, generating realistic adverse-condition variants at test time.

Various approaches have been proposed for providing synthetic AD data, to avoid the high costs of data collection and labelling.
These approaches include simulation \cite{dosovitskiy2017carla, nvidia2025drivesim, appliedintuition2025}, Domain Adaptation (DA) \cite{mullick2023daunit, li2024aldm} with GANs \cite{goodfellow2014gan} or Stable Diffusion (SD) \cite{rombach2022stable_diffusion}, Generative AI (GenAI) \cite{hu2023gaia1, wang2023drivedreamer}, and Neural Rendering \cite{yang2023unisim, zhou2024hugsim, yu2024sgd, chen2024omnire}.

\begin{figure*}
\vskip 0.2in
\begin{center}
    \centerline{\includegraphics[width=\textwidth]{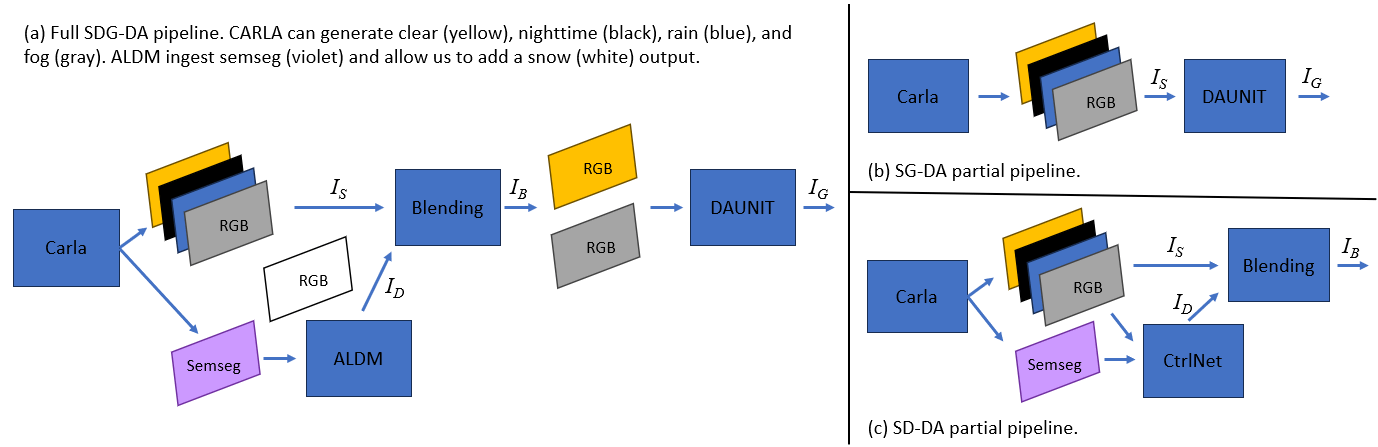}}

    \caption{
        \textbf{Data‑generation pipelines.}
        (a) \emph{SDG‑DA} (ours): CARLA provides clear/adverse renderings $I_S$ and semantic‑segmentation; ALDM converts the semseg into stylized image $I_D$, which we blend with $I_S$ to remove artifacts, yielding $I_B$. A DA‑UNIT model is trained to convert \emph{real} clear images into synthetic adverse images $I_G$.  
        (b) \emph{SG‑DA}: DA‑UNIT trained directly on CARLA pairs (w/o Diffusion).  
        (c) \emph{SD‑DA}: CARLA images $I_S$ pass through ControlNet to obtain $I_D$, which is blended with $I_S$ to give $I_B$ (w/o DA-UNIT).
    }
    
\label{fig_data_pipeline}
\end{center}
\vskip -0.2in
\end{figure*}

\begin{figure*}[t]
\vskip 0.2in
\begin{center}
    \centerline{\includegraphics[width=\textwidth]{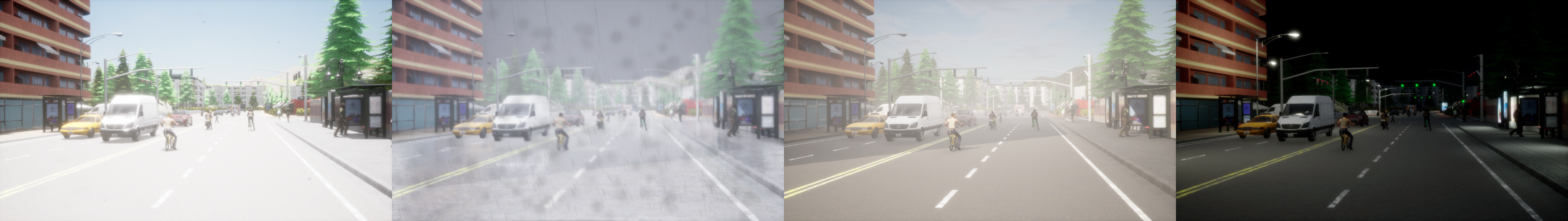}}
    \caption{
    \textbf{CARLA outputs.}
    Left to right: the original clear image, and the corresponding replayed images for rain, fog, and nighttime.
    }
\label{fig_carla_conds}
\end{center}
\vskip -0.2in
\end{figure*}

Here, we present a method to enhance a DA model by generating unique training data.
Our data generation pipeline is hybrid, using Simulation (S) and Diffusion (D), to train a GAN (G) on Clear-to-Adverse DA, hence we denote our solution as \textbf{SDG-DA}.
Going forward we use the notation in Figure ~\ref{fig_data_pipeline}, using $I_S$ for CARLA simulator outputs, $I_D$ for diffusion outputs, $I_B$ for the blending of $I_D$ with $I_S$, and $I_G$ for DAUNIT outputs.
In this work, the adverse conditions that we consider are fog, rain, snow, and nighttime.
We select DAUNIT \cite{mullick2023daunit} as our cornerstone for the high fidelity of its outputs to their corresponding ground-truth (GT) labels.
We start off by addressing DAUNIT's dependency on semantically matching clear-adverse image pairs, which are difficult to find in real-world AD datasets.
To resolve this challenge, we used the CARLA simulator \cite{dosovitskiy2017carla} to generate images ($I_S$) from a driving sequence under clear conditions, and develop the functionality to replay the sequence under adverse conditions (see Figure \ref{fig_carla_conds}).
Finally, motivated by prior research \cite{park2019spade, richter2021epe}, we modified the DAUNIT architecture to accept auxiliary inputs as priors, including semantic segmentation maps, instance segmentation masks, and depth maps.

To validate the utility of our synthesized adverse dataset, we train the REIN \cite{wei2024rein} semantic segmentation model on ACDC-Clear \cite{sakaridis2021acdc} plus our generated synthetic adverse images (without any real adverse images) and achieve $\mathbf{78.57\%}$ mIoU on the ACDC adverse test split. This result improves upon the pretrained REIN baseline and rivals state-of-the-art methods, all without using any real adverse data in training.

In the next sections we review related works, and elaborate on our contributions:

\begin{itemize}
  \setlength\itemsep{2pt}
  \setlength\topsep{2pt}
  \item We established a method to simulate matching clear–adverse pairs under multiple weather settings (Section \ref{methods_simulation}).
  \item We enhance photorealism and diversity with Stable Diffusion, while mitigating hallucinations via adaptive blending (Sections \ref{methods_diffusion}–\ref{methods_blending}).
    \item \textbf{Unified simulation $\rightarrow$ diffusion $\rightarrow$ GAN pipeline} that, for the first time, produces fully labelled \emph{synthetic} ACDC-Adverse images using only ACDC-Clear supervision (Sections~\ref{methods_simulation}–\ref{methods_daunit}).
    
    \item We extend DA-UNIT with auxiliary depth, semantic, and instance inputs, along with a clear–synthetic mixing schedule. We use object-detection ablation to identify the best data-generation setup, and demonstrate that this yields a 1.85\% relative mIoU improvement on ACDC-All (Sections \ref{methods_daunit}, \ref{results}).

\end{itemize}

\section{Related Work}\label{related_work}

\textbf{Unsupervised Domain Adaptation (UDA).}  
DAFormer/HRDA (56–65\% mIoU) \cite{hoyer2022daformer,hoyer2022hrda}, MIC (70\%) \cite{hoyer2023mic}, CoDA (72.6\%) \cite{hoyer2023hrda2}, and SoMA (78.8\%) \cite{yun2025soma} fine-tune on unlabelled ACDC-adverse (ACDC test scores shown in brackets).

\textbf{Domain Generalization (DG).}  
DG targets zero-shot performance on unseen domains (e.g.\ day~$\!\to\!$~night).  
The strongest scores on ACDC so far come from large vision–language backbones:  
VLTSeg \cite{vltseg} (77.9\% mIoU) fine-tunes EVA-02-CLIP ViT with Mask2Former.
REIN (77.6\%) attains comparable results while training less weights, by adding parameter-efficient adapters to the (frozen) backbone.
Both methods require {\em no} target-domain images but incur the overhead of a 300 M-parameter model.

\textbf{Image Restoration.}  
Image dehazing (RobustNet \cite{choi2021robustnet}, MWFormer \cite{zhu2024mwformer}) cleans the input before perception but adds latency and often fails under mixed weather conditions.

\textbf{Simulation.}  
Physical Based Rendering (PBR) engines such as CARLA synthesize multi-weather data, but photorealism, asset cost, and geographic coverage remain limiting.

\textbf{Generative Data Re-use.}  
Stable-Diffusion (SD) and ControlNet \cite{rombach2022stable_diffusion,zhang2023controlnet} enable text-prompt img2img translation but hallucinate in dense traffic scenes.  
Paired translation (UNIT / DAUNIT \cite{liu2018unitgan,mullick2023daunit}) improves fidelity, but requires real or simulated adverse-domain images.

\textbf{Our contribution (target-data-free data-generation).}
Using only the\\ \emph{ACDC-Clear} subset, we synthesize realistic adverse images via simulation~$\!\to$~diffusion~$\!\to$~GAN blending.
Fine-tuning a REIN backbone on this synthetic set lifts mIoU from \\75.1~$\to$~\textbf{75.5}\% on the adverse-only val split and from 75.1~$\to$~\textbf{76.5}\% on the full ACDC-All split.
Because the pipeline yields fully labelled images, the same data can be reused for \emph{any} vision task (detection, tracking, depth), whereas feature-level UDA or DG modifications benefit only segmentation.
Thus we provide an architecture-agnostic, \emph{task-agnostic} and cost-effective alternative to heavyweight foundation-model DG and data-hungry UDA.

\section{Proposed Method} \label{methods}

Our proposed data generation pipeline (SDG-DA) is composed of Simulation\\ (CARLA), Diffusion, and GANs (DAUNIT). For details and notation see Figure \ref{fig_data_pipeline}.

\subsection{Simulation} \label{methods_simulation}

We set out to generate matching clear-adverse image pairs for training DAUNIT.
We utilize CARLA's default weather parameters to simulate random driving scenarios under clear weather conditions; details can be found in our Supplementary Material (Section .1: CARLA Configuration, Table 1).
For each scenario, we parse the log file, containing events and states of all actors within the scene, namely vehicles and pedestrian poses. 
Next, we use the actor poses to replay the scenario under different weather conditions.
Ultimately, we obtain semantically matched simulated images under different environmental conditions (fog, rain, and nighttime)\footnote{Unreal Engine \cite{unrealengine2025} has plugins for snow simulation, these are not natively available in CARLA.}, which is practically impossible to accomplish on real sensor data.

\subsection{Diffusion} \label{methods_diffusion}

The CARLA simulated images $I_S$ tend to suffer from low photorealism, limiting their effectiveness. 
To bridge this sim2real gap, we post-process $I_S$ with the ALDM diffusion model \cite{li2024aldm}; its outputs are denoted as $I_D$ in Figure \ref{fig_data_pipeline}. ALDM parameters are detailed in Supplementary Material. One notable advantage of ALDM is its ability to generate realistic snow images, which is unavailable in CARLA. 
Additionally, ALDM improves certain visual aspects, such as the appearance of wet roads during rainy conditions.

\subsection{Blending} \label{methods_blending}

\begin{figure*}[t]
\centering
\includegraphics[width=1\linewidth]{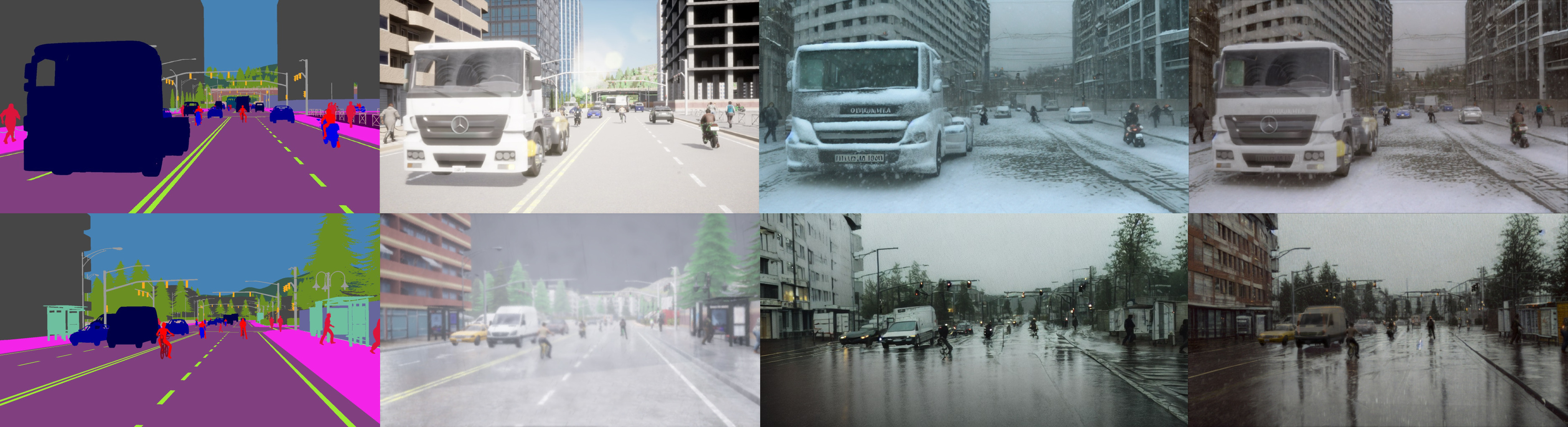}
\caption{
    \textbf{ALDM Blending.} Left to right: semseg, CARLA image $I_S$ (clear), ALDM output $I_D$ (top snow, bottom rain), blended result $I_B$.
}
\label{fig_aldm_blend}
\end{figure*}

Diffusion outputs, e.g. from ALDM, are prone to artifacts and hallucinations.
To address this, we estimate a quality map for each ALDM output $I_D$ and replace low-quality regions with the corresponding regions of the CARLA progenitor image $I_S$ (Figure \ref{fig_aldm_blend}).
While image composition is a well-established task \cite{niu2024makingimagesrealagain}, we opted for a pixel-wise weighted average of the CARLA and ALDM images, with spatially varying weights, as defined in \ref{eq_blending}:

\begin{equation}
\label{eq_blending}
I_{B} = w_{D} \cdot I_{D} + \left( 1 - w_{D} \right) \cdot I_{S}
\end{equation}

Here, $I_B$ is the blended output (size HxWx3), $I_S$ denotes the CARLA simulated image, $I_D$ represents the ALDM-generated output, and $w_D \in [0,1]$ is a pixel-wise blending weight map (size HxW, shared across channels).

The blending weights are determined using the semantic segmentation.
For example, ALDM struggles with bicycle riders so we assign them with low blending weights.
The weight map is then dithered and Gaussian-smoothed to ensure smooth transitions.

\subsubsection{Color Matching} \label{methods_colors}

ALDM ignores CARLA’s RGB statistics, so the simulated image \(I_S\) and its diffusion counterpart $I_D$ often have mismatched color palettes, producing visible seams during blending. We correct this with the Reinhard transformation (Eq.~\eqref{eq_colors}), applied first globally and then per object (using the instance‐segmentation map).
\begin{equation}
\label{eq_colors}
V' = \Bigl(\frac{\sigma_t}{\sigma_s}\Bigr)^{\gamma}(V_s - \mu_s) + \mu_t
\end{equation}
$V_s$ is the source image (mean $\mu_s$, std.\ $\sigma_s$), the target palette is denoted with \emph{t} ($\mu_t$, $\sigma_t$), and the strength is attenuated with exponent $\gamma$. Finally, to align the blended image \(I_B\) with the ACDC style, we sample one of 20 unlabelled calibration frames from the same weather slice and re‐apply Eq.~\eqref{eq_colors}. 
See implementation details in the Supplementary Material.

\subsection{DAUNIT} \label{methods_daunit}

\begin{figure*}[t]
\vskip 0.2in
\begin{center}
    \centerline{\includegraphics[width=\textwidth]{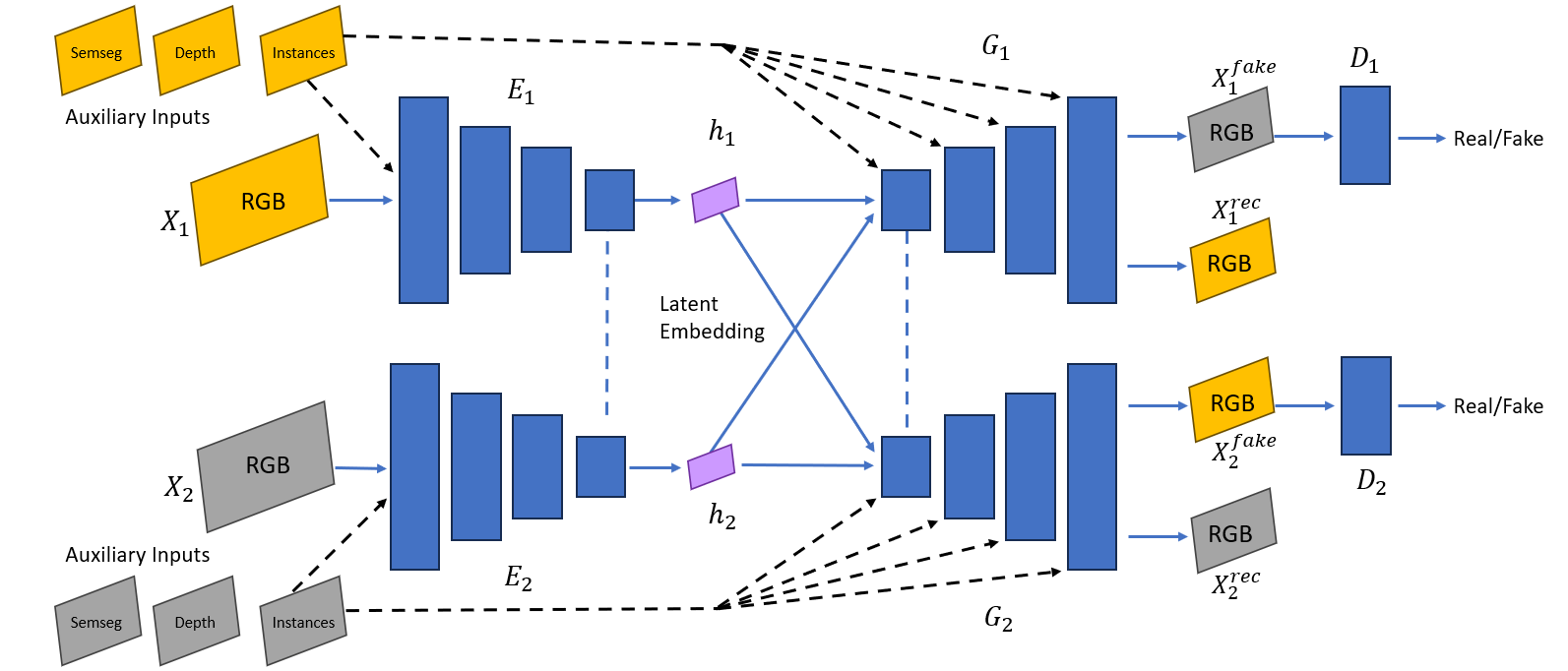}}
    \caption
    {
        DAUNIT architecture. 
        Encoders \(E_1,E_2\) map domain inputs \((X_1,X_2)\) to latents \(h_1,h_2\); generators \(G_1,G_2\) (weight-shared with the encoders, dashed) reconstruct or translate images, while discriminators \(D_1,D_2\) judge real/fake. 
        Optional auxiliary maps (semseg, instances, depth) can be injected at both encoding and decoding stages.
    }
    \label{fig_daunit_architecture}
\end{center}
\vskip -0.2in
\end{figure*}

\begin{figure*}[h]
\centering
\includegraphics[width=\linewidth]{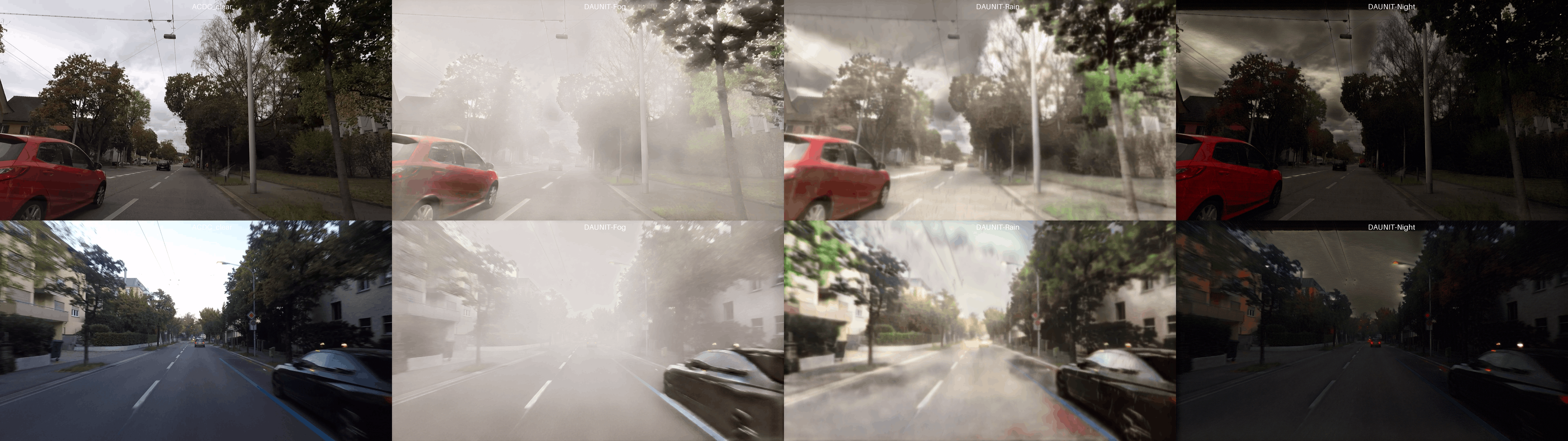}
\caption{
    \textbf{DAUNIT Outputs.}
    Left to right: clear ACDC input image and the corresponding fog, rain, and night outputs.
}
\label{fig_daunit_outputs}
\end{figure*}

As expected, our semantically aligned images from CARLA enhance DAUNIT outputs (Figure \ref{fig_daunit_outputs}).
However, we noticed a sim2real gap and hypothesized that mixing in real ACDC images while training DAUNIT, at a moderate ratio, could improve photorealism.
We refrained from using adverse ACDC images, to demonstrate our method's applicability to unseen\footnote{Excluding unlabelled color calibration images} target domains.
We found that randomly sampling clear real images with no semantically matching simulated counterparts is optimal at 10\%  (Table \ref{table_mixing}). Simulated images come with labelling, at no extra cost. At inference, we feed DAUNIT with real clear images from ACDC-Clear; the model then generates corresponding synthetic adverse images (fog, rain, snow, or night) for downstream perception tasks.
Following \cite{park2019spade, richter2021epe}, we extend DAUNIT to ingest auxiliary inputs, namely depth, semantic segmentation, and instance segmentation, in both the encoder and decoder (Fig.~\ref{fig_daunit_architecture}). See Section 9 of the Supplementary Material for details.

\subsection{Benchmarking on Alternative Data Sources} \label{methods_benchmarking}

To benchmark our effectiveness, we consider alternative data sources for training.

\textbf{Image augmentation} could be applied as an alternative DA method, for example Foggy Cityscapes \cite{sakaridis2018foggy}.
Image augmentation may refer to various non-learnable techniques, including gamma corrections, color shifts, blurring, and overlaying.
These methods are often fast enough to be applied on-the-fly during training, but might require some expertise to configure correctly.
We benchmarked our pipeline against corresponding image augmentations and provide implementation details in the Supplementary Material.


\textbf{ACDC-All} includes real adverse images, which provide novel objects and semantic layouts but incur additional collection and labelling costs. It is not a standard benchmark; we include it for cost-effectiveness analysis and ablation.

\section{Experiments and Results}\label{results}

We fine-tuned downstream tasks on ACDC-Clear + our synthetic adverse images and evaluated on ACDC validation splits:

\textbf{Object Detection.}  
      Serves as an ablation probe: training on candidate synthetic pipeline configurations, and ranking by Average Precision (AP) on ACDC to select the most effective.

\textbf{Semantic Segmentation.}  
We evaluate the improvement over the REIN baseline using mean Intersection over Union (mIoU), reporting results on the \emph{Adverse-only} slice (Fog, Rain, Snow, Night) and on the full \emph{ACDC-All} set (Adverse + Clear).

\subsection{Object Detection} \label{res_obj_det}

In this task, we expanded our benchmarking to include a small ablation study on our pipeline components.
First, we replace ALDM \cite{li2024aldm} with ControlNet (CN) \cite{zhang2023controlnet}, and to distinguish between these two variations we denote them in Table \ref{table_objdet_ap_results} as SGD-DA ALDM and SGD-DA CN.
Next, we remove DAUNIT and examine the partial pipeline denoted as SD-DA in Figure \ref{fig_data_pipeline}.
To avoid diffusion artifacts we implemented SD-DA using the more conservative ControlNet.
Furthermore, we expanded our blending method to take in more inputs and be more stringent.


We fine-tuned EVA-02 \cite{eva02} on various datasets, and evaluated it on ACDC according to the COCO protocol \cite{coco2023eval}.
Each experiment was repeated five times, and in Table \ref{table_objdet_ap_results} we show the mean Average Precision (AP) over repetitions.
The EVA-02 model was pre-trained on COCO; 
Each training dataset included a copy of ACDC-Clear.

Notably, the highest AP scores were achieved by SG-DA, without diffusion.
An observant reader will notice the unexpectedly high fog scores; both this anomaly and a credit-assignment ablation that quantifies each weather condition’s contribution are detailed in the Supplementary Material.

\begin{table*}[t]
 \footnotesize                    
  \setlength{\tabcolsep}{4pt}      
  \renewcommand{\arraystretch}{0.95}
\caption
{
    Object-detection average precision (AP, \%) on the ACDC val splits. 
    Columns correspond to individual weather conditions; the best score in each column is \textbf{bold}. 
    “ACDC-All” is trained on all weather conditions and serves as an upper-bound, so it is not bolded.
}
\label{table_objdet_ap_results}
\vskip 0.15in
\begin{center}
\begin{small}
\begin{sc}
\begin{tabular}{lccccccc}
    \toprule
    Method & Fog & Rain & Night & Snow & Clear & Adverse & All \\
    \midrule
    ImgAug          & \textbf{49.60} & 31.47 & 22.00 & 37.28 & 31.23 & 33.27 & 32.51 \\
    SG-DA              & 47.55 & \textbf{32.82} & \textbf{22.05} & 38.25 & 31.28 & \textbf{33.85} & \textbf{33.02} \\
    SGD-DA CN        & 46.81 & 32.03 & 21.45 & \textbf{39.17} & 31.22 & 32.83 & 32.26 \\
    SGD-DA ALDM      & 45.67 & 31.23 & 20.84 & 37.34 & \textbf{31.71} & 32.45 & 32.20 \\
    \midrule
    ACDC-All          & 53.02 & 36.14 & 25.59 & 47.02 & 32.31 & 39.05 & 36.43 \\
    \bottomrule
\end{tabular}
\end{sc}
\end{small}
\end{center}
\vskip -0.1in
\end{table*}

\subsection{Semantic Segmentation} \label{res_semseg}

We fine-tune REIN, pretrained on Cityscapes and fine-tuned on ACDC-Clear as our baseline, whereas prior work typically uses Cityscapes-only models. We report mean Intersection over Union (mIoU) on the ACDC validation split and observe significant gains over this strong baseline: +1.39\,mIoU (+1.9\%) overall and +2.67\,mIoU (+4.6\%) at night (Table~\ref{talbe_semseg_scores}). Each experiment is repeated ten times and significance is confirmed via the Wilcoxon signed-rank test ($p<0.1$).

For consistency with object detection, every training set includes ACDC-Clear. From Section~\ref{res_obj_det}, we combine ImgAug, SD-DA (CN) and SG-DA into our final dataset. Clear weather sees the largest boost (+3.15\,mIoU, +4.0\%), outperforming ACDC-All by +2.90 mIoU (+3.7\%), likely due to a 5:1 ratio of synthetic to adverse images.

\begin{table*}[t]
 \footnotesize                    
  \setlength{\tabcolsep}{4pt}      
  \renewcommand{\arraystretch}{0.95}
\caption{
    Semantic-segmentation mean IoU (mIoU, \%) on the ACDC val splits.
    Columns correspond to the per-weather slices; the best score in each column is \textbf{bold}.
    “ACDC-All” is trained on all weather conditions and acts as an upper-bound, so it is not bolded.}
\label{talbe_semseg_scores}
\vskip 0.15in
\begin{center}
\begin{small}
\begin{sc}
\begin{tabular}{lcccccc}
\toprule
Method & Fog & Rain & Night & Snow & Clear & All \\
\midrule
ACDC-Clear & \textbf{83.48} & 78.65 & 57.81 & 76.97 & 78.57 & 75.09 \\
ImgAug     & 83.32 & \textbf{79.51} & \textbf{60.61} & 77.79 & 78.20 & 75.86 \\
Ours       & 83.35 & 78.98 & 60.48 & \textbf{77.85} & \textbf{81.72} & \textbf{76.48} \\
\midrule
ACDC-All   & 85.53 & 82.35 & 62.12 & 82.54 & 78.82 & 78.27 \\
\bottomrule
\end{tabular}
\end{sc}
\end{small}
\end{center}
\vskip -0.1in
\end{table*}

\subsection{Ablation Studies} \label{ablations}

\subsubsection{Synthetic Data Pipeline Components}

We gauged each component by retraining EVA-02 on ablated pipeline variants (Table \ref{table_pipeline_justification}). The ALDM and CN variants use raw diffusion outputs \(I_D\), skipping our blending stage. ACDC-Clear fine-tuning is the baseline in every synthetic set. All ablations reduce AP, while a strong image-augmentation baseline boosts it. These results confirm that both DAUNIT and Diffusion-CARLA blending are essential and that simulation plus diffusion alone cannot match their impact.

\begin{table}[t]
 \footnotesize                    
  \setlength{\tabcolsep}{4pt}      
  \renewcommand{\arraystretch}{0.95}
\caption{
    \textbf{Pipeline components impact.}
    AP scores on ACDC-All validation split.
    CARLA simulated data $I_S$, even w/ Diffusion post-processing ($I_B$), is inferior w/o DA-UNIT.
}
\label{table_pipeline_justification}
\vskip 0.15in
\begin{center}
\begin{small}
\begin{sc}
\begin{tabular}{lcc}
\toprule
Trained On     & AP    \\
\midrule
    $I_S$ & 29.89 \\ 
    $I_S$ \& $I_B$ (ALDM) & 20.61 \\
    $I_S$ \& $I_B$ (CN) & 29.46 \\
    ImgAug & \textbf{32.51} \\
\bottomrule
\end{tabular}
\end{sc}
\end{small}
\end{center}
\vskip -0.1in
\end{table}

\subsubsection{Auxiliary Inputs}

The major modification we introduce to DAUNIT is auxiliary inputs, according to the principles in EPE \cite{richter2021epe}.
We compare different modalities: depth, semantic segmentation (semseg), and instance segmentation (inst).
We begin with a minimalist approach, providing auxiliary inputs only at the encoder input layer.
Next, we inject these auxiliary inputs after the model's bottle neck, in addition to providing them at the encoder input layer.
More explicitly, this is done at multiple resolutions into the encoder (Figure \ref{fig_daunit_architecture}).
The AP differences in Table \ref{table_aux} span only $\pm$1.75\% from the vanilla model.
In proceeding experiments we used the optimal configuration, i.e. only depth at the input layer.
Visualization of the output differences can be found in the Supplementary Material.

\begin{table}[t]
  \centering
  \footnotesize                    
  \setlength{\tabcolsep}{4pt}      
  \renewcommand{\arraystretch}{0.95}
  \caption{
    \textbf{Auxiliary inputs.}
    Impact of adding Depth, Semseg, and Instance maps to RGB in DAUNIT’s encoder (and optionally decoder).}
  \label{table_aux}
  \begin{tabular}{ccccc}
    \toprule
    Depth & Semseg & Inst & Dec. & AP \\
    \midrule
     - & - & - & - & 32.45 \\
     x & - & - & - & \textbf{33.02} \\
     - & x & - & - & 32.95 \\
     - & - & x & - & 32.85 \\
     x & x & - & - & 32.07 \\
     x & - & x & - & 31.99 \\
     - & x & x & - & 32.34 \\
     x & x & x & - & 32.05 \\
     x & - & - & x & 32.18 \\
     - & x & - & x & 32.75 \\
     - & - & x & x & 32.29 \\
     x & x & - & x & 32.58 \\
     x & - & x & x & 31.90 \\
     - & x & x & x & 32.89 \\
     x & x & x & x & 32.14 \\
    \bottomrule
  \end{tabular}
  \vskip -0.1in
\end{table}


\subsubsection{Synthetic Real Data Mixing}

When training DAUNIT on blended inputs, we observed a sim2real gap. To address this, we augment our training set with real source-domain images (i.e., ACDC-Clear). Specifically, in each batch we randomly replace a certain percentage of the synthetic clear images with real ACDC-Clear images. As Table \ref{table_mixing} demonstrates, low mixing ratio yields meaningful performance gains.

\begin{table}
 \footnotesize                    
  \setlength{\tabcolsep}{4pt}      
  \renewcommand{\arraystretch}{0.95}
\caption{
    \textbf{Mixing Ratios.}
    AP scores for different proportions of random real images (ACDC-Clear) mixed during training.
}
\label{table_mixing}
\vskip 0.15in
\begin{center}
\begin{small}
\begin{sc}
\begin{tabular}{cc}
\toprule
    Mixing Ratio & AP               \\
    \bottomrule
    0            & 32.66            \\
    10           & \textbf{33.02}   \\
    30           & 32.30            \\
    \bottomrule
\end{tabular}
\end{sc}
\end{small}
\end{center}
\vskip -0.1in
\end{table}

\vspace{-4pt}

\section{Discussion}

In this work, we address the challenge of limited adverse-condition data by proposing a suite of diverse generation methods for training perception algorithms. We provide a thorough analysis of each approach’s advantages and limitations, backed by comprehensive empirical evidence. Our experiments highlight two key benefits: (i) blending each diffusion output with its source image, and (ii) enhancing DAUNIT with matched simulation pairs and auxiliary inputs.
Notably, we train DAUNIT almost entirely on CARLA-simulated clear–adverse pairs (with only a 10\% ACDC-Clear mix), and at test time feed it real ACDC-Clear images to synthesize realistic adverse-weather variants.

Our best object-detection results come from the SG-DA partial pipeline (Figure~1). For segmentation, using our final dataset (including ImgAug), we achieve a \textbf{+1.39 mIoU (1.9\%)} overall improvement over the ACDC-Clear baseline when evaluated on ACDC-All, closing 42\% of the gap to the ACDC-All upper bound. Notably, our method also outperforms ACDC-All model by \textbf{+2.90 mIoU (3.7\%)} in the clear weather condition.

Our method reaches \textbf{78.57\% mIoU on ACDC-Adverse (test)}, approaching \\SoMA (78.8\%) \cite{yun2025soma} without depending on real adverse data. The data is fully labeled and task-agnostic, making it suitable for object detection and depth estimation and beyond.

Finally, the per-condition performance in ACDC-All mirrors real-world accident rates, supporting the claim that perception failures under adverse conditions are a major hazard.
To facilitate qualitative inspection, we provide an interactive demo showcasing our domain adaptation pipeline and its results at: \href{https://yahiashowgan.github.io/bridging-driving-conditions/}{yahiashowgan.github.io/bridging-driving-conditions}.

\clearpage
\newpage
\newpage

\appendix

\section*{Appendix}
\addcontentsline{toc}{section}{Appendix}

This document provides additional implementation details, configurations, and ablation studies to support the results presented in our paper titled \emph{“Bridging Clear and Adverse Driving Conditions: Domain Adaptation with Simulation, Diffusion, and GANs”}.

The sections below are organized according to the structure of the main paper, and aim to facilitate reproducibility and clarity.

\subsection{CARLA Configuration} \label{app_carla_config}

To minimize redundancy, we set a low frame rate (0.05 FPS) and retained only 15 frames per scenario. 
The probability of spawning a pedestrian, a two-wheel platform (bicycle, motorcycle) and a four-wheel platform (car, bus, truck) were set to be equal.
In comparison to typical real-world datasets (\ref{app_acdc_obj_dist}), our dataset is intentionally skewed towards the pedestrian and Vulnerable Road User (VRU) categories, both considered to be more challenging for AD perception systems.

For each weather condition we sample uniformly the value of each parameter according to \ref{tbl_weather}. 
The scattering parameters were held constant and shared across all conditions.

\begin{table*}[hb]
\caption{CARLA Weather Parameters.}
\label{tbl_weather}
\vskip 0.15in
\begin{center}
\begin{small}
\begin{sc}
\resizebox{\textwidth}{!}{%
\begin{tabular}{lcccr}
    \toprule
    Parameter & Clear & Fog & Rain & Night \\
    \midrule
    cloudiness & 0.0 - 30.0 & 10.0 - 60.0 & 30.0 - 90.0 & 0.0 - 40.0  \\
    dust storm & 10.0 - 50.0 & 0.0 - 20.0 & 0.0 - 20.0 & 10.0 - 50.0  \\
    fog density & 0.0 - 0.1 & 20.0 - 40.0 & 0.0 - 7.0 & 5.0 - 15.0  \\
    fog distance & 300.0 - 1000.0 & 7.0 - 20.0 & 6.0 - 10.0 & 3.0 - 100.0  \\
    fog falloff & 0.1 - 0.2 & 1.0 - 4.0 & 0.1 - 0.5 & 0.1 - 1.0  \\
    precipitation & 0.0 - 0.1 & 0.0 - 7.0 & 60.0 - 100.0 & 0.0 - 0.1  \\
    precipitation deposits & 0.0 - 0.1 & 10.0 - 30.0 & 50.0 - 90.0 & 0.0 - 20.0  \\
    sun altitude angle & 30.0 - 90.0 & 30.0 - 90.0 & 30.0 - 90.0 & -90.0 - -45.0  \\
    sun azimuth angle & 0.0 - 360.0 & 0.0 - 360.0 & 0.0 - 360.0 & 0.0 - 360.0  \\
    wetness & 0.0 - 10.0 & 60.0 - 100.0 & 0.0 - 40.0 & 0.0 - 60.0  \\
    wind intensity & 0.0 - 20.0 & 0.0 - 10.0 & 30.0 - 100.0 & 0.0 - 20.0  \\
    MIE scattering scale & 0.03 & 0.03 & 0.03 & 0.03  \\
    Rayleigh scattering scale & 0.0331 & 0.0331 & 0.0331 & 0.0331  \\
    scattering intensity & 1.0 & 1.0 & 1.0 & 1.0  \\
    \bottomrule
\end{tabular}
}
\end{sc}
\end{small}
\end{center}
\vskip -0.1in
\end{table*}

Until recently, CARLA used Unreal Engine 4 \cite{unrealengine2025}, which suffered from limited photorealism.
This is expected to improve with Unreal Engine 5, which was unavailable at the time of this project.

\subsection{ALDM Parameters} \label{app_aldm_params}

We apply ALDM using 25 inference steps and an unconditional guidance scale of 7.5.
To increase the diversity and quantity of our dataset, each simulated semseg map is processed three times using different random seed values.
To generate different weather conditions we provide ALDM the following prompts:
\begin{itemize}
\setlength\itemsep{1pt}
  \item Clear: sunny day.
  \item Fog: super heavy fog.
  \item Rain: rainy scene, heavy rain, wet roads, lights reflecting on road.
  \item Night: nighttime, extremely dark.
  \item Snow: snowy scene, lots of snowflakes blowing.
\end{itemize}

\subsection{Color Blending} \label{app_color_blending}

We introduced the power parameter $\gamma$ in Section 3.3.1 in the paper in the following Eq.:
\begin{equation}
\label{eq_colors_2}
V' = \Bigl(\tfrac{\sigma_t}{\sigma_s}\Bigr)^{\gamma}(V_s-\mu_s)+\mu_t
\end{equation}
After searching different values, we selected 0.5, i.e. taking the square root.
We applied \ref{eq_colors_2} in the Lab color space, and added custom handling of out of range values.
For each channel we check if the values exceed the upper limit, in which case we scale all values above $\mu_t$ to [$\mu$, upper limit].
The same is applied to values below $\mu$, bringing all values into the valid range while retaining the target mean value.

When matching ALDM to ACDC, we applied a secondary blending between the color-matched image $I^{\prime}_{ALDM}$ and the original ALDM image $I_{ALDM}$, according to \ref{eq_acdc_blend}.

\begin{equation}
\label{eq_acdc_blend}
I^{\prime\prime}_{ALDM} = w_{orig} \cdot I_{ALDM} + \left( 1 - w_{orig} \right) \cdot I^{\prime}_{ALDM}
\end{equation}

The parameter $w_{orig}$ is uniformly sampled from the range [0, 0.5]. 
This secondary blending allows us to retain some more of the diversity introduced by ALDM while aligning with ACDC color palette.

\subsection{ALDM Blending Parameters} \label{app_aldm_blend}

ALDM outputs were given the following blending weight, according to the semantic labels

\begin{itemize}
\setlength\itemsep{1pt}
    \item 0.9: unlabeled, sidewalk, building, wall, fence, vegetation, terrain, sky, static, other, water, ground, bridge, guard rail.
    \item 0.8: road.
    \item 0.7: pole, dynamic, road line, rail track.
    \item 0.5: traffic sign.
    \item 0.3: pedestrian, rider, motorcycle, bicycle.
    \item 0.1: traffic light, car, truck, bus, train.
\end{itemize}

\subsection{Alternative Blending Methods} \label{app_alt_blend}

In \cite{niu2024makingimagesrealagain} image composition is broken down into several sub tasks:

\begin{itemize}
\setlength\itemsep{1pt}
    \item Object placement.
    \item Semantic appearance variation.
    \item Image blending, i.e. avoiding unnatural boundaries.
    \item Image harmonization, ensuring that the object lighting is compatible to the background lighting.
    \item Shadow generation.
    \item Reflection generation.
\end{itemize}

Object placement is taken care for during simulation, and passing its outputs to an image-to-image diffusion model preserves correct placement.
ALDM \cite{li2024aldm} used the semantic segmentation (semseg) map, and we also experimented with ControlNet \cite{zhang2023controlnet}, in which case we tried different combinations of control inputs, including semseg, instance segmentation, depth, and edges.
The diffusion process increases appearance variation, especially when using different random seeds.

Blending of boundaries is handled by our color matching and quality weighted pixel wise averaging of the CARLA and ALDM images.
We tried using Poisson Image Editing (PIE) to blend in objects, which matches the object's boundary to the background, and then searches for a function that is compatible with the object's Laplacian.
In our experience PIE worked poorly, both the standard implementation and the optimized version \cite{fast_green_pie}, perhaps due to the complexity of the driving images.

For harmonization we tried applying PCT-Net \cite{pctnet} which worked well, but we came to the conclusion that the effort was not justified for two main reasons.
Evaluation analysis indicated that we should focus on small remote objects, on which lighting is less of an issue and harmonization barely modifies them.
Moreover, PCT-Net runtime was very long, since driving images can contain dozens and even hundreds of objects, and each object has to be treated separately.
We attempted to speed up PCT-Net by aggregating all the inserted objects and handling all of them as a single object, but this led to image corruption.

Shadows are taken care for during the CARLA simulation.
Reflections are most pronounced on wet roads during rain, and ALDM handles reflections quite well.
This was actually one of our decision factors to use ALDM and not ControlNet.

For our trials with ControlNet, we employed an alternative implementation of our blending concept. Further details will be disclosedin the future.

\subsection{Image Augmentation} \label{app_imgaug}

We applied image augmentations based on onto clear ACDC images to obtain four adverse conditions.
Our implementations are based on Albumentations \cite{albumentations} and Kornia \cite{kornia}.

\subsubsection{Fog}

\begin{itemize}
\setlength\itemsep{1pt}
    \item Color de-saturation by 40\%.
    \item Color shift to blue, by additional 20\% de-saturation of the red and green channels.
    \item Additive fog overlay, implemented as semi transparent ($\alpha = 0.3$) 2000 randomly distributed 2D Gaussians.
\end{itemize}

\subsubsection{Rain}

\begin{itemize}
\setlength\itemsep{1pt}
    \item Color de-saturation by 60\%.
    \item Color shift to blue, by additional 40\% de-saturation of the red channel and 20\% de-saturation of the green channel.
    \item Reducing image brightness linearly. 
    \item Reducing image brightness with a gamma correction, $\gamma=1.5$.
    \item Additive rain drops, implemented as up to three collections of short semi-transparent ($\alpha = 0.2-0.3$) lines at similar slant angles. Each collection containing 100-500 line elements, with similar color and length.
    \item Additive windshield drops, implemented as a fog overlay but configured to have\\ smaller radii and sharper definition.
    \item Gaussian blurring.
    \item Glass blurring, also known as pixel-shuffle, within a small radius (4 pixels).
\end{itemize}

\subsubsection{Snow}

\begin{itemize}
\setlength\itemsep{1pt}
    \item Over exposing bright regions to mimic snow deposits, using albumentation's \\RandomSnow class with \textit{"bleach"} method.
    \item Additive snow flakes, using albumentation's RandomSnow class with \textit{"texture"}\\ method.
    \item Color de-saturation by 70\%.
    \item Color shift to blue, by additional 40\% de-saturation of the red and green channels.
    \item Reducing image brightness linearly. 
    \item Reducing image brightness with a gamma correction, $\gamma=2.0$.
\end{itemize}

\subsubsection{Night}

\begin{itemize}
\setlength\itemsep{1pt}
    \item Darkening the sky by additive fog on the top half of the image.
    \item Color de-saturation by 40\%.
    \item Color shift away from blue, implemented with a color mixing matrix configured to reduce the blue channel by 10\% while transferring 50\% of its energy to the red and green channels.
    \item Reducing image brightness linearly. 
    \item Reducing image brightness with a gamma correction, $\gamma=2.3$.
\end{itemize}

\begin{figure*}[h]
    \centering
    \includegraphics[width=1\linewidth]{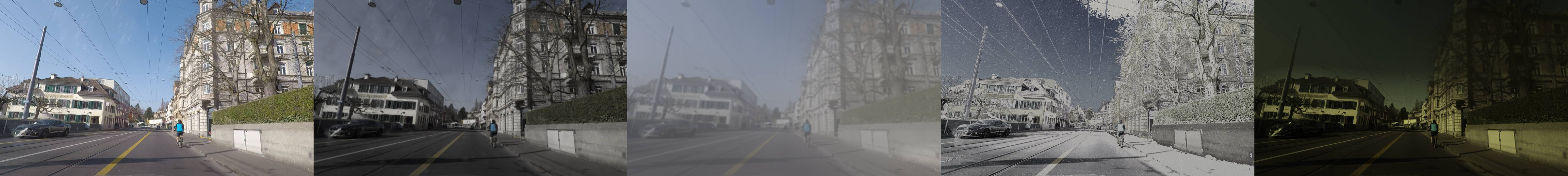}
    \includegraphics[width=1\linewidth]{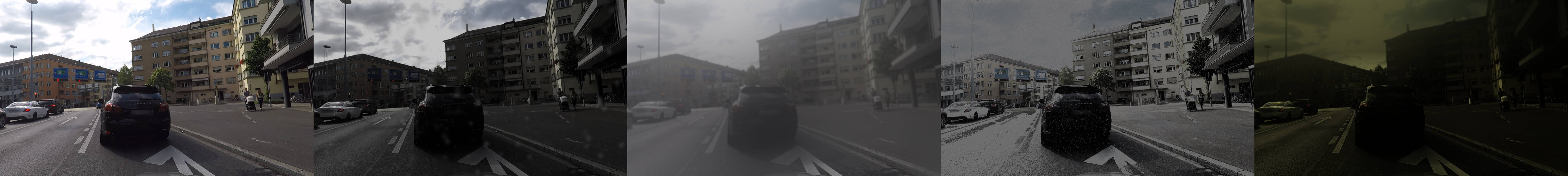}
    \includegraphics[width=1\linewidth]{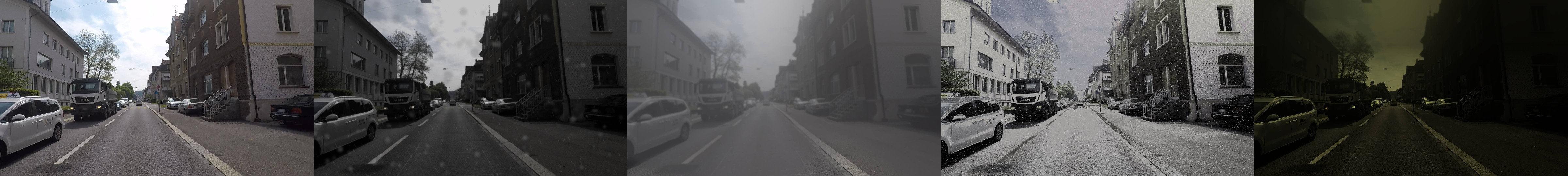}
    \caption{
        \textbf{Image Augmentation examples.}
        Left to right: clear ACDC input image, augmented outptus for rain, fog, snow, and night.
    }
    \label{fig_app_imgaug}
\end{figure*}

\subsection{Weather Conditions Credit Assignment Study} \label{app_credit}

We conducted a credit assignment study by training on data from a single condition and evaluating across all data slices, adhering to the ACDC train/val splits.
Results for training on real images (ACDC-Adverse) are presented in \ref{table_credit_assignment}.
As expected, the best training data for each condition was its own. However, fog benefited more from training on snow and rain than from fog-specific data.

\begin{table*}[h]
\caption{
    \textbf{Real data credit assignment.}
    Performance of models trained on ACDC-clear with individual weather splits (rows) evaluated across different ACDC weather conditions (columns). 
    Each cell represents the corresponding metric score.
    Bolding denotes best score column-wise, excluding the training on ACDC-All.
}
\label{table_credit_assignment}
\vskip -0.15in
\begin{center}
\begin{small}
\begin{sc}
\begin{adjustbox}{max width=\textwidth}

\begin{tabular}{lccccccc}
    \toprule
    Trained On & Fog & Rain & Night & Snow & Clear & Adverse & All \\
    \midrule
    ACDC-clear + Fog   & \cellcolor{gray!20}44.60 & \cellcolor{gray!20}31.86 & \cellcolor{gray!20}21.76 & \cellcolor{gray!20}39.78 & \textbf{32.31} & 33.69 & 33.28 \\
    ACDC-clear + Rain  & \cellcolor{gray!20}51.06 & \cellcolor{gray!20}\textbf{35.86} & \cellcolor{gray!20}21.53 & \cellcolor{gray!20}40.96 & 31.68 & \textbf{36.29} & \textbf{34.61} \\
    ACDC-clear + Night & \cellcolor{gray!20}46.15 & \cellcolor{gray!20}32.36 & \cellcolor{gray!20}\textbf{24.50} &\cellcolor{gray!20} 37.44 & 32.02 & 34.09 & 33.51 \\
    ACDC-clear + Snow &	\cellcolor{gray!20}\textbf{51.35} &	\cellcolor{gray!20}34.18 &	\cellcolor{gray!20}21.78 &	\cellcolor{gray!20}\textbf{44.48} &	31.86 &	35.33 &	33.73
    
    \\
    ACDC-clear + All  & 53.02	&	36.14	&	25.59	&	47.02	&	32.31	&	39.05	&	36.43\\
    
    \bottomrule
\end{tabular}
\end{adjustbox}
\end{sc}
\end{small}
\end{center}
\vskip -0.1in
\end{table*}

Repeating the same study on our synthetic data reveals a more complex pattern (\ref{table_credit_assignment_daunit}).
Overall, synthetically generated night and rain data provide the greatest benefits. Our training-validation scores rule out over-fitting as a plausible explanation.
However, combining all three synthetic conditions does not yield the best results (not shown).

\begin{table*}[h]
\caption{
    \textbf{Synthetic data credit assignment (DAUNIT).}
    Performance of models trained on our generated dataset splits (rows) evaluated across ACDC weather conditions (columns).
    Each cell represents the corresponding metric score.
    Bolding denotes best score column-wise, excluding the training on ACDC-All.
}
\label{table_credit_assignment_daunit}
\vskip -0.15in
\begin{center}
\begin{small}
\begin{sc}
\begin{adjustbox}{max width=\textwidth}
\begin{tabular}{lccccccc}
    \toprule
    Trained On & Fog & Rain & Night & Snow & Clear & Adverse & All \\
    \midrule
    ACDC-Clear + DAUNIT-fog   & \cellcolor{gray!20}46.45 & \cellcolor{gray!20}\textbf{32.13} & \cellcolor{gray!20} 20.34 & 37.38 & 31.50 & 32.69 & 32.28 \\
    ACDC-clear + DAUNIT-rain  & \cellcolor{gray!20}47.15 & \cellcolor{gray!20}31.85 & \cellcolor{gray!20}\textbf{21.79} & 38.15 & 31.84 & \textbf{33.67} & 33.09 \\
    ACDC-clear + DAUNIT-night & \cellcolor{gray!20}\textbf{48.40} & \cellcolor{gray!20}31.34 & \cellcolor{gray!20}21.34 & \textbf{38.36} & \textbf{32.45} & 33.52 & \textbf{33.25} \\
    ACDC-clear + DAUNIT-all   & 47.54 & 32.66 & 20.80 & 38.44 & 31.96 & 33.11 & 32.58 \\
    \bottomrule
\end{tabular}
\end{adjustbox}
\end{sc}
\end{small}
\end{center}
\vskip -0.1in
\end{table*}

\subsection{Object Distributions and Difficulty of Different Weather Conditions}\label{app_acdc_obj_dist}

\begin{figure} [h]
\centering
\includegraphics[width=0.5\linewidth]{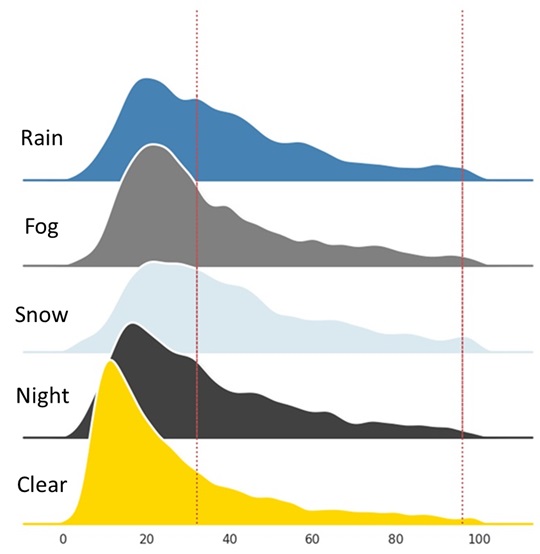}
\caption{
    \textbf{Object Size Distribution.} 
    Objects in adverse weather are larger than in clear conditions. The x-axis represents object size (equivalent radius). The red lines denote COCO's thresholds for small, medium, and large objects.
}
\label{fig_app_obj_size}
\end{figure}

We observed that some adverse condition splits in ACDC yielded higher scores than evaluations on ACDC-Clear (see Section 4.1 and Table 1 in the main paper).
To understand why object detection performed worse on clear images than on adverse images, we analyzed object distribution across different weather conditions.
First, we depict the size distribution of objects in each weather condition in \ref{fig_app_obj_size}.
Clear conditions contain the highest proportion of small objects, and the average object size is significantly smaller than in adverse conditions.

\begin{figure} [h]
\centering
\includegraphics[width=1\linewidth]{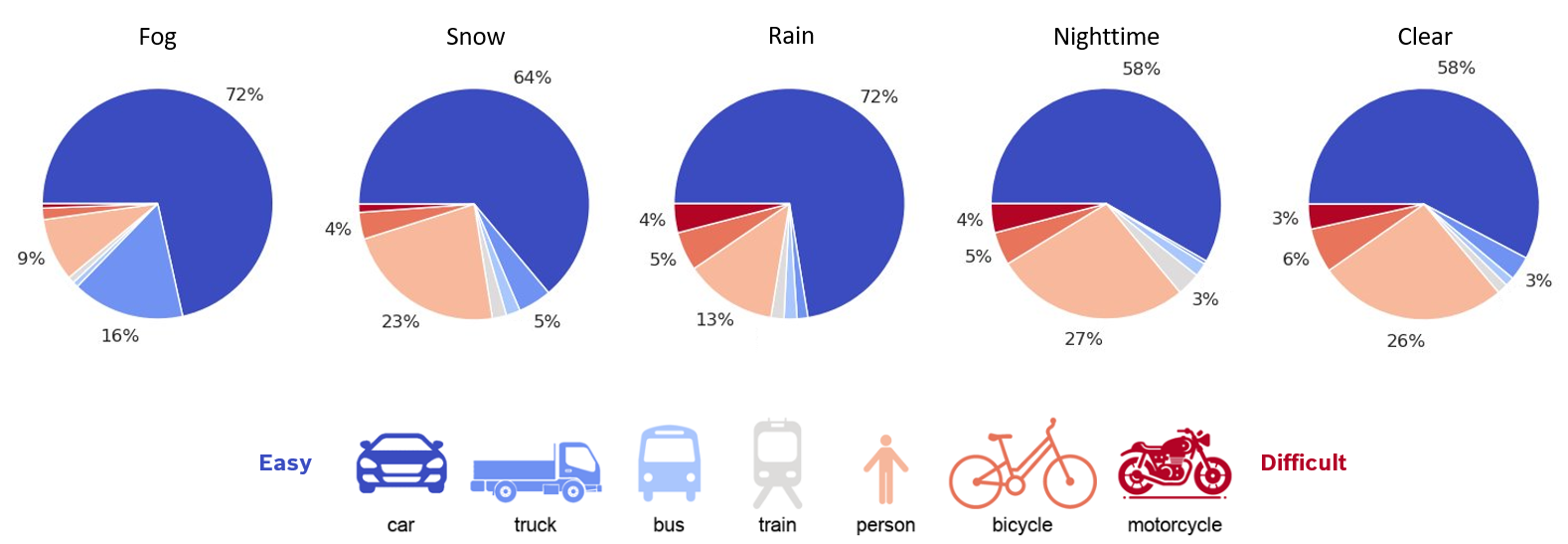}
\caption{
    \textbf{Object Category Distribution} 
    Clear and nighttime conditions contain much more difficult objects than fog.
    Objects difficulty is ranked by EVA-02 Average Precision (AP) scores.
}
\label{fig_app_obj_category}
\end{figure}

This effect is further compounded by the fact that more difficult objects appear more frequently in certain weather conditions than others. 
To illustrate this, we analyze the object distribution across different weather conditions in ACDC (see \ref{fig_app_obj_category}).

On average, the ACDC-Clear split contains more difficult and smaller objects compared to adverse conditions.
This difference in object distribution likely contributes to the lower AP scores observed for ACDC-Clear, despite its more favorable weather conditions.
While adverse weather conditions introduce visibility challenges, they often contain fewer and larger objects, which are easier to detect, leading to higher AP scores in certain cases.

\subsection{Additional DAUNIT Configuration Parameters} \label{app_ablation_daunit}

The first DAUNIT configuration parameter we explored is the reconstruction loss.
The original loss function was based on a distance metric computed on VGG \cite{vgg} embeddings.
In \ref{table_losses}, we compare this to newer loss functions: distance on ConvNeXt \cite{convnext} embeddings and LPIPS \cite{lpips} perceptual distance.
All models were trained on CARLA images with a 10\% mix of ACDC-Clear, using both Patch and Multi-scale discriminators.
The only difference between the models was the choice of reconstruction loss.

\begin{table} [h]  
    \centering
    \begin{minipage}{\linewidth} 
        \centering
        \begin{tabular}{lc}
            \toprule
            Loss       & AP   \\
            \midrule
            VGG        & 32.36 \\
            ConvNeXt   & 31.81 \\
            LPIPS      & \textbf{33.02} \\
            \bottomrule
        \end{tabular}
    \end{minipage}
    \caption{
        \textbf{DAUNIT Reconstruction Loss Comparison.} 
        Average Precision (AP) results for different reconstruction loss functions used in DAUNIT training.
    }
    \label{table_losses}
\end{table}

Next, we quantify the impact of training DAUNIT on matched images.
We hypothesize that the Patch and Multi-scale discriminators respond differently, prompting us to evaluate their combinations (\ref{table_discriminators}).

Using both discriminators together consistently yielded the best results for object detection.
Visually, adaptation to nighttime was most effective with the Patch discriminator, while adaptation to fog and rain looked better with the Multi-scale discriminator and matched images.

We incorporated depth as an auxiliary input at the encoder stage, employed LPIPS as the reconstruction loss, and utilized 10\% clear data mixing in all training configurations.
The only variation in the setup was the type of discriminator used.

\paragraph{Patch Discriminator vs. Multi-Scale Discriminator}

Patch and multi-scale discriminators are used in Generative Adversarial Networks to improve the realism of generated images, but they evaluate images differently.

A patch discriminator divides an image into small patches and evaluates each one separately instead of classifying the entire image as real or fake.
This approach encourages the generator to focus on local textures and fine details, making it useful for preserving spatial structures in image-to-image translation.

A multi-scale discriminator uses multiple discriminators at different resolutions, analyzing an image from a global low-resolution view to a local high-resolution perspective.
This setup helps improve both overall structure and fine-grained details, making it effective for domain adaptation and tasks requiring consistency across different scales.

Visually, the Multi-scale discriminator produced more realistic images for rain and fog, while the Patch discriminator performed better for nighttime.
However, for training EVA in object detection, image matching had no significant impact, regardless of whether the Patch or Multi-scale discriminator was used.

\begin{table}[h]
\caption{
    \textbf{Discriminator and Image Matching.}
    Average Precision (AP) on ACDC-All for different DAUNIT configurations. 
    “Matching” denotes whether semantically matched image pairs were used during training.
}
\label{table_discriminators}
\vskip 0.15in
\centering
\begin{small}
\begin{sc}
\begin{tabular}{cc|c|c}
    \toprule
    \multicolumn{2}{c|}{\textbf{Discriminator Type}} & \textbf{Matching} & \textbf{AP} \\
    \cmidrule(lr){1-2}
    \textbf{Multi-scale} & \textbf{Patch} & & \\
    \midrule
    \checkmark & -- & -- & 31.84 \\
    \checkmark & -- & \checkmark & 31.88 \\
    -- & \checkmark & -- & 32.69 \\
    -- & \checkmark & \checkmark & 32.58 \\
    \checkmark & \checkmark & \checkmark & \textbf{33.02} \\
    \bottomrule
\end{tabular}
\end{sc}
\end{small}
\vskip -0.1in
\end{table}

\subsubsection{DAUNIT Auxiliary Inputs} \label{app_daunit_aux}

The auxiliary DAUNIT inputs are pre-processed as follows: 

\begin{itemize}
    \item \textbf{Depth}: Depth was normalized to the [0, 1] range. 
    It was generated using MiDaS \cite{Midas}, a monocular depth estimation method.
    \item \textbf{Semantic Segmentation}: Following EPE \cite{richter2021epe}, semantic segmentation was provided as binary maps.
    Instead of using a single channel with \(N\) enum values, we provide a tensor of \(N\) channels with binary values.
    \item \textbf{Instance Segmentation}:
    Each image has a different number of object instances, which can vary greatly, from single-digit to the hundreds of objects.
    To regularize the instance segmentation representation, we calculate the instance adjacency matrix and apply greedy coloring \cite{greedy_color}.
    This yields instance segmentation maps with five distinct values at most. 
\end{itemize}

To see the difference in outputs of DAUNIT using different types of aux input, take a look at Table \ref{fig_app_aux_inputs}:

\newcolumntype{Y}{>{\centering\arraybackslash}X} 


\begin{table}[H]
  \centering
  \setlength{\tabcolsep}{2pt}
  \begin{tabularx}{\columnwidth}{YYY}
    \bfseries ACDC Input 
      & \bfseries Semseg \& Instance (Encoder)
      & \bfseries Depth (Encoder) \\[3pt]
    \multicolumn{3}{c}{%
      \includegraphics[width=0.99\linewidth]{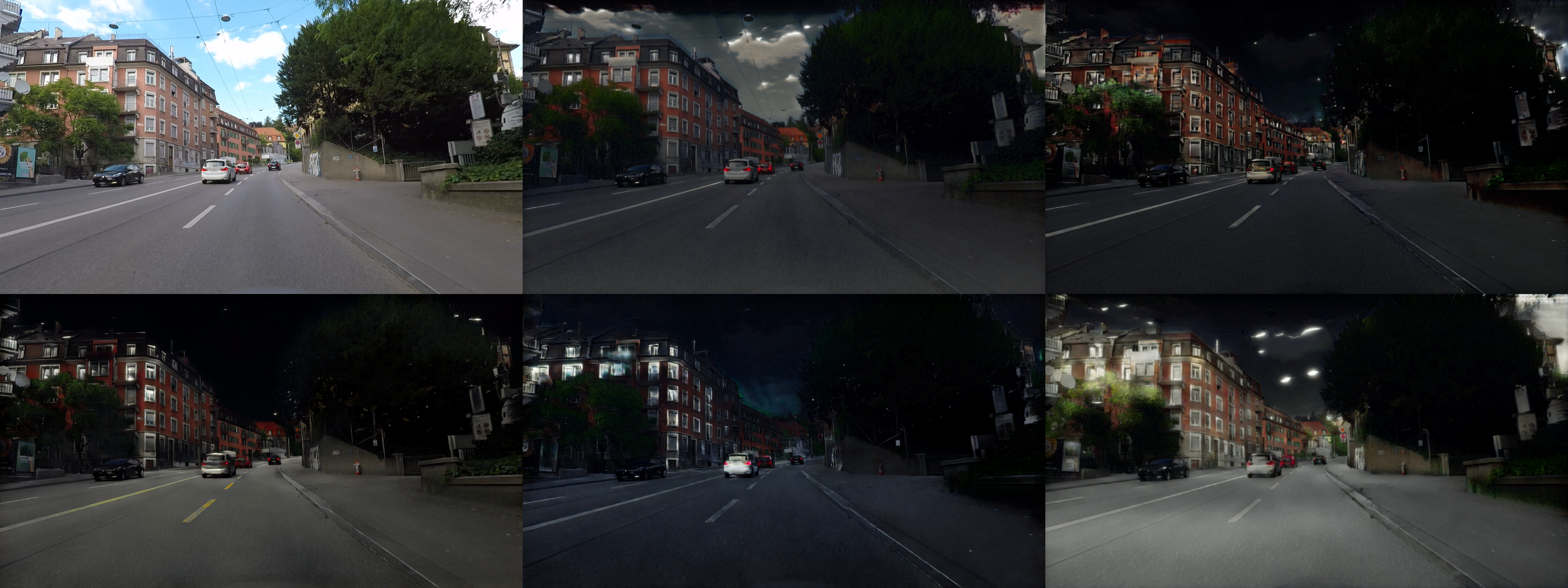}%
    } \\[3pt]
    \bfseries No aux.\ input
      &\bfseries Semseg \& Instance (Enc.\ \& Dec.)
      &\bfseries Depth (Enc.\ \& Dec.) \\
  \end{tabularx}
  \caption{\textbf{Auxiliary Inputs Configuration.}
    (a) ACDC input. (b)–(f) DAUNIT outputs under different aux.\ maps.}
  \label{fig_app_aux_inputs}
\end{table}

\bibliographystyle{plain}
\bibliography{egbib}

\end{document}